**World Scientific**
www.worldscientific.com

# FUSION BASED HAND GEOMETRY RECOGNITION USING DEMPSTER-SHAFER THEORY

Asish Bera, Debotosh Bhattacharjee , Mita Nasipuri

Department of Computer Science and Engineering, Haldia Institute of Technology,

Haldia, WB, India

Department of Computer Science and Engineering, Jadavpur University,

Kolkata-700032, India.

e-mail: asish.bera@gmail.com. debotosh@ieee.org, mitanasipuri@gmail.com

This paper presents a new technique for person recognition based on the fusion of hand geometric features of both the hands without any pose restrictions. All the features are extracted from normalized left and right hand images. Fusion is applied at feature level and also at decision level. Two probability based algorithms are proposed for classification. The first algorithm computes the maximum probability for nearest three neighbors. The second algorithm determines the maximum probability of the number of matched features with respect to a thresholding on distances. Based on these two highest probabilities initial decisions are made. The final decision is considered according to the highest probability as calculated by the Dempster-Shafer theory of evidence. Depending on the various combinations of the initial decisions, three schemes are experimented with 201 subjects for identification and verification. The correct identification rate found to be 99.5%, and the False Acceptance Rate (FAR) of 0.625% has been found during verification.

## 1. Introduction

**D**ata fusion is generally used to attain the best optimal solution by combining two or more different solutions of a problem.[1] It has already widely been accepted in the various application domains, like remote sensing, robotics, weather forecasting, etc. The 'Biometric' is another pertinent field where fusion has been applied at a large scale. Biometric traits are the appropriate substitute of the conventional knowledge-based (password) or token-based (PIN) user verification systems in various required levels (low to high level) of security intelligence. [4] It is applied as one of the best reliable and legitimate human authentication systems in a constrained environment. The primary objective is to discriminate the identity of a person based on various unique biometric properties recognized by an automated system. These systems are developed by the physical (e.g. face, fingerprint, hand geometry, hand vein etc.) and behavioral (e.g. gait, signature, voice etc.) distinctiveness of an individual.[6] Different human organs are employed individually (unimodal) or combined (multimodal) for this purpose. A standalone recognition decision by a unimodal system is not always reliable and robust enough to verify whether a person is genuine or imposter. So to enhance the performance, fusion[5] is very useful to authenticate an individual with multibiometrics. It basically combines several decisions taken by various expert systems. Multibiometric systems provide certain benefits over the limitations of unimodal systems such as noise-effect, intra-class variations, non-universality, spoof-attack, and flexibility.[4] Different multibiometric systems are accessible and are classified according to their basic properties (e.g. multi-sensor) and level(s) of implementation (e.g. decision level). An enormous number of biometric systems with different characteristics and functionalities are running successfully worldwide for several decades in the government (National ID cards), forensics (criminal investigations) and commercial applications (smart cards).[1]

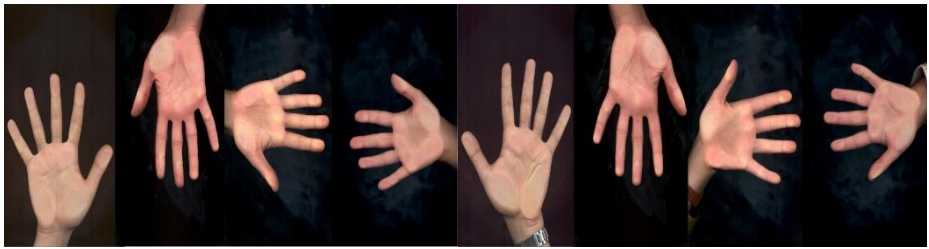

Fig.1. Sample left and right hand images at different pose.

Hand geometry is one of the oldest physiological biometric techniques, commercially started its journey with the *'Identimat'* in early 1970s that pioneered other biometric devices .[13] It is used for verification in various kinds of access control systems. It is advantageous due to lower implementation cost, smaller template size, and easier accessibility.[7] It also bears certain drawbacks, mainly:

(i) larger device dimension precludes embedding in smaller devices, like laptop and Smartphone.

(ii) contact between hand and device may cause hygienic problems.

(iii) lack of higher uniqueness, universality, and acceptability.

(iv) finger sizes and associated measurements may change due to age or environmental factors during image acquisition. Fingernail is also a significant factor to locate finger extremes and feature extraction.

Initially, hand biometric systems  had been using a fixed contact-based rigid pattern, called 'peg'; and CCD cameras or scanners were utilized for image acquisition.[14] Afterwards, the peg-based systems have been substituted by peg-free or contact-free systems, providing more user flexibility during image collection.[8] The weaknesses of utilizing traditional CCD camera have been overcome by the expensive infrared camera.[16, 18, 27] Research attention is also concentrating on 3D hand geometry[12] and combination of 2D and 3D hand features.[13] The feature set mainly consists of  the measurements of palm width and palm area, finger lengths and widths at different positions on the fingers. In literature, generally the number of such geometric features varies from 15 to 40. The number of snapshots taken from left and/or right hand may vary from 3 to 10 per subject. Other than considering all fingers or the entire hand, Zhou *et al.* has experimented particular finger geometry. [20] For higher accuracy, multimodal systems are developed using fusion at various levels. In multimodal hand geometry, most of the techniques are integrated with other hand-based traits (e.g. fingerprint, palmprint, vein pattern etc.). In the case of non-hand based modes, face modality has been considered. But, multiple traits mostly require high resolution camera or scanner which is expensive and the environmental factors (such as weather, lighting conditions etc.) need to be considered during image acquisition. Dissimilar feature extraction techniques may increase design complexities due to heterogeneous feature set and size of the database. Some of the relevant state-of-the-arts are depicted in Table1.

Performance of Near Infrared (NIR) hand images are investigated in recent years. NIR images have been experimented with 20 subjects (30 features/user) by Support Vector Machine (SVM).[27] In another work [16], thermal images are used, and the system has been developed using extension theory with 34 features for 30 persons (10 images per subject). Fusion of 2D and 3D hand features has been experimented with 177 subjects, and it supports fusion of heterogeneous (hand geometry, finger texture and palmprint) feature set. [13] Fusion of 2D and 3D face and hand features are tested with 15 people.[12] Hand geometry and palmprint has been introduced with fusion at matching score level.[17] Score level fusion with various normalization techniques (e.g. min-max, z-score etc.) are presented to 100 individuals.[6] Decision level fusion using majority voting, AND, OR rule with 86 users (10 images per user) has been implemented. [25] Hand shape based approach using Independent Component Analysis (ICA1and ICA2) has been presented with458 right hand subjects.[9]

In this present work, hand mode has been used with low cost scanner to minimize computational complexities in terms of feature definition and classification algorithms. The primary objective of this work is to improve the performance of a hand biometric system using data fusion at the feature and decision level. The methods for image normalization, locating landmark points, and feature extraction from the normalized hand images are exploited. Image normalization is defined to characterize uniform hand template that involves segmentation, rotation, and largest component detection. Two probabilistic classification algorithms are presented based on which different experiments are carried out for identification. Finally, persons are classified by the Dempster-Shafer[2, 3] belief functions. The major novelty incorporated in this system is the application of the belief function at decision level fusion. The experiments are performed based on the fusion of hand features extracted from both the hands of a subject.

Rest of this paper is organized as follows: Section 2 illustrates various fusion strategies developed in reference to hand biometrics. The system model of the present work is outlined in Section 3. Section 4 defines the Dempster-Shafer theory of evidence briefly. Section 5 describes various steps required to normalize the original color hand images with two major issues. Section 6 depicts the extraction of all the geometric features after normalization. Section 7 is presented with the classification algorithms. Section 8 discusses the experimental results, followed by the conclusion in Section 9.

Table 1. Study of some fusion based hand biometric systems.

| Author, Year [Ref.] | Modes[*] | Feature Definition | Subject (images.) | Fusion Level |
|---|---|---|---|---|
| Ribaric & Fratric, 2005[10] | HG+PP | Eigenfinger+eigenpalm(byKLT) | 237(5&10) | Score & Decision |
| Kumar & Zhang, 2006[11] | HG+PP | 23 HGFs+DCT of PP | 100(10) | Feature |
| Choraś &Choraś, 2006[17] | HG+PP | 8 region areas+ZMI of PP | 100(3) | Score |
| Hanmandlu *et al.*, 2008[24] | HG+PP | 23 HGFs+144 DCT of PP | 100(10) | Decision(by PSO) |
| Ross *et al.*,2005[22] | HG+F | 9-byte HGFs+LDA & PCA of face | 100(5) | Feature |
| Strintzis *et al.*, 2007 [12] | HG+F | 2D &3D of HGFs+"face-ness" | 50(70) | Score (2D and 3D) |
| Jain *et al.*, 2005 [6] | HG+FP+F | 14 HGFs+minutiae+Eigenface | 50(5) | Score |
| Alonso *et al.*, 2007 [21] | HG+PP+FT | 15HGFs+texture of PP &FT | 109(10 | Score, Feature &Decision |
| Zhou *et al.*, 2007 [26] | HG+PP+FP | 16HGFs+wavelets of PP+FP | 98(10) | Feature and Score |
| Vasirkala *et al.*, 2010 [23] | HG+PP+HV | ICA+gabor wavelets of PP &HV | 100(3) | Score(t-norm) |
| Zhang *et al.*, 2011[13] | HG+FT+PP | 25HGFs+CompCode+SurfaceCode | 177(10) | Score (2D and 3D) |
| Shahin *et al.*,2011[18] | DHG+FP | 29HGFs+minutiae | 100(10) | Score |

[*]Few abbreviations [HG=Hand Geometry; PP=Palmprint; KLT=Karhunen-Loeve (K-L) Transform; HGF= Hand Geometric Feature; DCT=Discrete Cosine Transform; ZMI= Zernike Moment Invariants; PSO= Particle Swarm Optimization; F= Face; LDA=Linear Discriminant Analysis; PCA= Principal Component Analysis; FP=Fingerprint; FT= Finger Texture; HV= Hand Vein; ICA= Independent Component Analysis; DHG=Dorsal Hand Geometry].

## 2. Related works on fusion in hand biometrics

The preliminary discussion about the classification of multimodal hand biometrics and its levels of fusion is investigated exhaustively. Some recent and relevant works are also sorted out briefly.

### 2.1. *Multibiometrics*

A biometric system operates in four different modules: sensor, feature extraction, matching, and decision making module.[1,4,5,6,7] At every module fusion can be applied. Fusion is divided as pre-matching and post-matching.[6] Pre-matching fusion is used at the sensor and feature extraction module. Post-matching fusion is implemented in the matching and decision module. According to the basic properties, multibiometrics can be categorized into the following five types. [1]

(i)**Multi-sensor:** two or more sensors are employed for image acquisition from same or different traits with better accuracy at higher implementation cost.[6]

(ii)**Multi-modal:** images from several modalities (e.g. HG+F) are collected using same or different sensors to produce better performance. Processing complexity is higher due to heterogeneous feature extraction and matching techniques. Table 1 depicts some works on multi-modal biometric systems.

(iii)**Multi-sample:** variations (e.g. hand pose and accessories) of a selective trait from different types of samples are integrated to achieve precise and optimal representation of the user identity. [1]

(iv)**Multi-instance:** images from two or more instances (left and right hand) of a particular mode are collected from a person to extract more robust features, and that causes larger size of the database. [8, 29]

(v)**Multi-algorithm:** more than one algorithm is applied for feature extraction (e.g. hand shape and texture features) or matching at higher computational complexities. [8]

The palmprint (PP) is a better conjugate of hand geometry (HG) [30]. It is cost efficient due to the hand-based mode. The face (F), fingerprint (FP) and other modes are fused with hand geometry to achieve better performance. Hand vein (HV) pattern matching is a promising area that can be combined with HG. Finger knuckle print (FKP) [29] and dorsal hand geometry (DHG) [18] are also being investigated.

## 2.2. *Levels of fusion*

Fusion method is classified according to various levels as applied either at the pre-matching or post-matching, from the raw data to final decision level. These levels are as follows:

i) **Sensor level** fusion is achieved in multi-sensor and multimodal biometrics when more than one sensor is employed, such as 2D or 3D imaging devices.[6]

ii) **Feature level** fusion is applicable on same or different types of feature set.[22] Mean value is used for compatible features. Template concatenation technique or another combination rule such as a statistical approach, like Canonical Correlation Analysis (CCA) is applied to incompatible features.[20]

iii) **Score level** fusion involves the computation and integration of similarity measures. Sometimes, the score is normalized, and some popular methods used are: sum rule, weighted sum rule, product rule, min/max rule, and z-score rule, etc. [6, 23]

iv)**Rank level** fusion consolidates the matching scores by assigning ranks to the identities according to order of matching confidence from multiple matchers. It contains lesser information about the identities and provides better security in multimodal systems. The Borda count, the highest rank, and Bucklin majority voting are traditional methods. [28]

v) **Decision level** fusion combines decisions taken by the individual matcher independently to reach final decision. Threshold is determined by using the Euclidean, Absolute, Hamming or Mahalanobis distance metrics. The AND/OR rule is a famous one [24], and the majority of the voting algorithm has been implemented by Li *et al.*[25]

It is clear from Table1 that score level and decision level have been explored reasonably, and feature level is given more attention. The rank level is not exploited yet using hand geometry. A palmprint based rank level fusion has been proposed by Kumar *et al.*[28]

## 3. **System model**

The generalized system model is depicted in Fig.2. Preprocessing contains all the steps required for normalization. After extracting features from both the hands of a test subject, they are matched against the same enrolled hand template database by two classifiers. The feature templates are fused after calculating the Euclidean distance differences, and subsequent probabilities are computed. Then, first level decisions are made by the classifiers according to the maximum probabilities and are denoted by Decision1 and Decision 2. The final decision is taken according to the maximum belief function, whether to accept or reject a claimed identity at decision level. The significant properties of this system are:

(i) single sensor is required and unimodal system,

(ii) probabilistic and multi-algorithm based, and

(iii) works on a similar database.

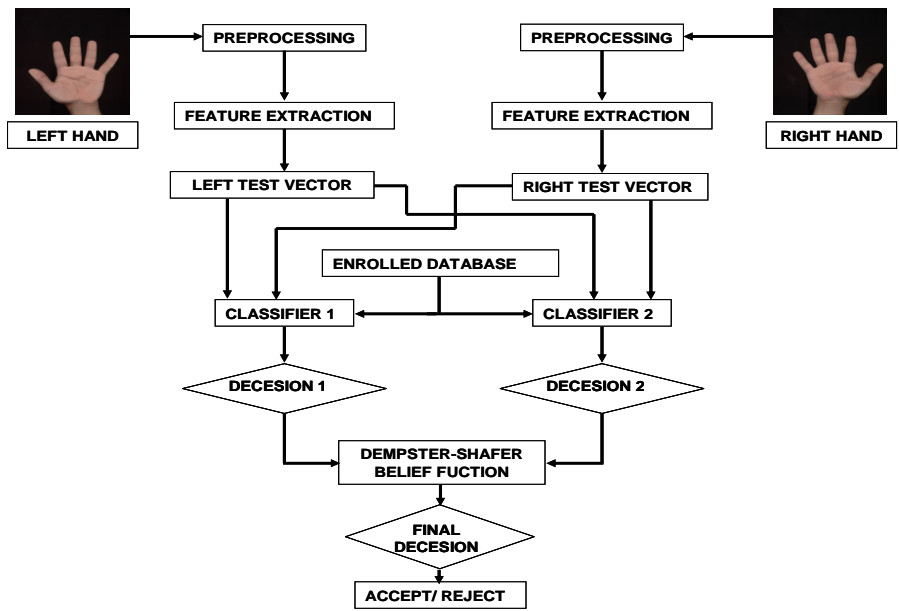

Fig.2. Fusion based hand biometric system.

Three different techniques are experimented using this underlying system model. The main difference is found in the first level decision. In all experiments, for initial decision making three classes are considered. But, number of classes is extendable according to the classification algorithms by increasing the value of the nearest neighbors. The recognized classes are either one each from left, right and fused feature templates or all from the fused feature set, as defined in the experiment section. The resultant classes from the two classifiers are matched, and their BPAs are assigned accordingly at the final decision making stage. A class with the maximum probability is delineated as the last decision. In coming few sections, detailed explanations of every module are specified.

## 4. Dempster-Shafer theory of evidence

The original evidence theory was introduced by A. P. Dempster in 1960s to generalize the Bayesian framework and modified by G. Shafer in 1976. The Dempster-Shafer theory (DST) [2, 3] is used to produce a belief function based on a combination rule. The basic combination rule is described here concisely. Let, $U$ be the set of mutually exclusive and exhaustive hypotheses of a problem, known as frame of discernment and $2^U$ be the set of all subsets of $U$ in the interval [0,1], including the null set (Ø) and full set. The basic probability assignment function $m(*)$ can be defined as:

$$\begin{cases} m(\phi) = 0 \\ \sum_{A \subseteq U} m(A) = 1 \end{cases}$$

(1)

The belief (BEL) and plausibility (PL) functions of $A$ can be given as:

$$BEL(A) = \sum_{A \subseteq U} m(A) \quad \text{and} \quad PL(A) = 1 - BEL(\bar{A})$$

(2)

where $BEL(A)$ and $PL(A)$ represent the lower bound and upper bound of belief in $A$. If $m_1$ and $m_2$ represents two Basic Probability Assignments (BPAs) from independent evidences, and the condition $\sum_{\substack{i,j \\ A_i \cap B_j = \phi}} m_1(A_i) m_2(B_j) < 1$ is satisfied, then the combined BPA can be defined as:

$$\begin{cases} m(C) = \dfrac{\sum_{\substack{i,j \\ A_i \cap B_j = C}} m_1(A_i) m_2(B_j)}{1 - \sum_{\substack{i,j \\ A_i \cap B_j = \phi}} m_1(A_i) m_2(B_j)} \\ m(\phi) = 0 \end{cases}$$

(3)

The combination of $m_1$ and $m_2$ is possible if and only if there exist at least two subsets $A_i$ and $B_j$ with $A_i \cap B_j = \phi$ such that $m_1(A_i) \neq 0$ and $m_2(B_j) \neq 0$. The combination rule is extendable for several belief functions. The basic combination rule (Eq.3) has been modeled in this scheme at decision level by generating the BPAs by two classification methods independently. The belief functions are defined for three variables. Detailed descriptions are given in Section 7.3.

## 5. Hand image normalization

Hand image normalization is the most significant step to define a precise hand template, and the system performance is heavily reliant on this normalization. This process consists of several successive sub steps. These are specifically background elimination, rotation, removal of wrist irregularities, and fingertip-valley localization. Two important issues are also discussed that may create a problem during normalization.

## 5.1. *Hand segmentation*

The original color images consist of actual hand texture with or without any stuff as foreground and dark background. Conversion from a color image into a grayscale image is performed using thresholding by *Otsu's* method[15] and median filtering is applied for noise removal. Then, the grayscale image is converted into binary form, and the hand contour is traced using the *'Sobel's* edge detection method, in Fig.3 (b).

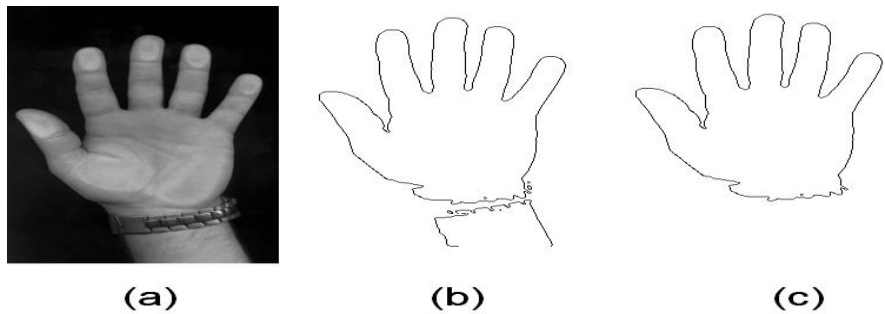

**(a)**                    **(b)**                    **(c)**

Fig.3. (a) grayscale image (b) contour detection (c) main component detection.

## 5.2. *Rotation and component detection*

The sample images had acquired at different posture, placed at various angles (Fig.1) with a little consideration that no two fingers are being overlapped or touched. To define uniform template, a particular orientation for all hand contour is employed by rotating that counter clockwise at various angels (e.g. 90,180 degree), with respect to its centroid. The centroid $(\bar{x}, \bar{y})$ is defined as, $(\bar{x}, \bar{y}) = (M_{10}/M_{00}, M_{01}/M_{00})$.                    (4)
where, $M_{i,j}$ is the moment of the binary image. Rotation is useful to make every hand contour perpendicular with the X-axis by aligning straight line passing through center of the contour to Y-axis. But, after rotation the hand shape may be distorted and pixel intensities may change at the new location. Some of the pixels may not have corresponding pixel in original hand image. Therefore, bilinear interpolation has been used to circumvent shape deviation after rotation. Due to ornament, wristwatch, ring or any other hand gadget wore by a person during image acquisition, the obtained binary image may be broken into several components, which is shown in Fig.3(b). The largest component detected, which is of prime interest for this work, is kept intact and rest smaller components are eliminated. In the wrist region of the central component, some geometrical irregularities exist which are removed by smoothing. This type of unpredictable irregularities changes from person to person who may degrade the recognition rate. To solve this problem, a reference line (UV) at a distance $h$ is considered whose lower portion should contain the rough wrist region. The value of $h$ is determined as the 20% of the total length of the main component. The leftmost and rightmost nonzero pixels are the two end points of UV. The lower portion of line UV is neglected (Fig.4). The midpoint of line UV is delineated as the reference point R, which is important to locate the finger tips and valleys.

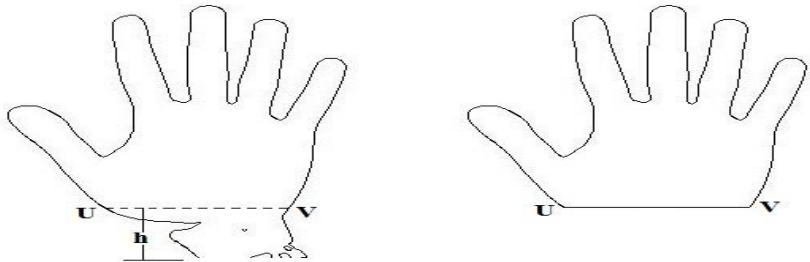

Fig.4. (a) Removal of wrist irregularities (b) Resultant hand after smoothening.

### 5.3. *Finger tips and valleys localization*

Fingertip and valley points are localized by a new procedure that has been applied to both kinds of hands. The method proposed by Yörük *et al.*[9] for finding extreme points of fingers from the hand contour used radial distance with respect to a reference point around the wrist region. But, the basis of the radial function is not mentioned correctly. Here, the entire hand contour is divided logically into left and right parts with respect to the line RG formed between the reference point (R) and the middle finger tip, denoted by G, which is defined by the maximum Euclidean distance from R. In case of left hand, the left half contains the tips of thumb and index fingers along with their associated valleys. The right half consists of the tips and valleys of ring and little fingers of the same hand. For the right hand tips and valleys, the reverse scenario takes place. To trace the tips and valleys two parameters are considered: (i) maximum Euclidean distances from R and (ii) associated angles formed with the reference line UV and line produced between R and respective finger extreme points. The localization of tips and valleys is defined using these parameters. The sequence of locating the tip-valley points of a left hand is: middle tip (G), thumb tip (B), first thumb valley (C), index tip (E), middle-index valley (F), little tip (K), first little valley (J), ring tip (I), and ring-middle valley (H), as shown in Fig.5(b). The second valley of thumb (A), ring (D), and little (L) fingers are defined on the other side of finger contour by calculating same distance between the fingertip and the first valley of the corresponding finger.

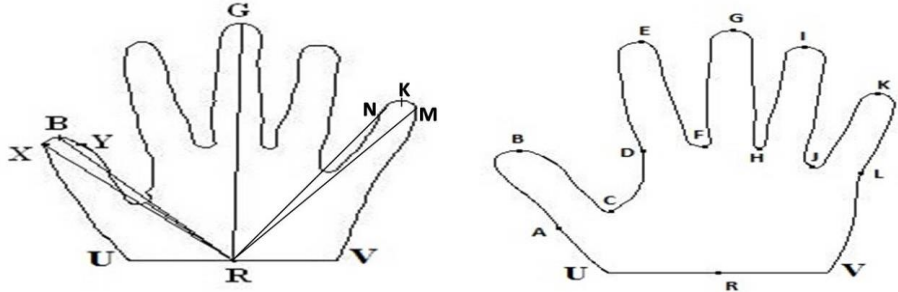

Fig.5. (a) tip localization of thumb and little finger, (b) all finger tips and valleys localization.

First of all, the maximum Euclidean distance from point R$(x_1, y_1)$ is considered to find a middle finger tip point G$(x_2, y_2)$. The angle (inverse tangent in degree at counter clockwise direction) formed between UV and RG is calculated as:

$$\theta = tan^{-1}((y_2 - y_1)/(x_2 - x_1)), \tag{5}$$

where, $(x,y)$ represents the pixel of point R and G. Then, find out the leftmost (for left hand) point X on the contour to trace temporary thumb-tip. Scan the pixels to the right side from point X horizontally (row wise) to detect the first intersection point (Y) on the hand contour. Between the angle $\angle XRY$ (generated by line RX and RY with respect to R) the maximum Euclidean distance from R is located as the actual thumb tip B. The minimum distance from R within angle $\angle BRG \pm \alpha^\circ$ represents the first valley point (C) of thumb. The tolerance angle $\alpha^\circ$ (here it is fixed to $\pm 5^\circ$) is considered due to pose variations and to yield more accuracy in this procedure for all the remaining fingers. The maximum distant point within $\angle CRG \pm \alpha^\circ$ is represented as index finger tip (E). Valley F is marked by considering the point with minimum distance within $\angle ERG \pm \alpha^\circ$. Similar procedure is followed for the right side of this logical partitioning of hand contour to locate the tip-valley points of little and ring fingers. To find little finger tip (K), the rightmost nonzero pixel (M) is identified. Then, horizontal scanning towards the left side is performed to find another pixel (N) on the little finger contour, and the corresponding angle $\angle MRN$ is measured. Finally, the maximum distant point within $\angle MRN$ from R is fixed as little finger tip K. Then, according to similar steps mentioned above, the valley (J), tip (I), and valley (H) are marked by evaluating angles and highest distance from R within the range of angles. At this stage, only one valley point from each of the thumb, index and little fingers are located. Distances from the identified tip to valley of these corresponding fingers are calculated. The unidentified valley points are traced by measuring equal distances from the finger-tips to locate other valley points of the respective fingers. Say, for thumb the distance between B and C is measured and then find the point (A) that is the same distance apart from B on the other side of thumb. All the identified points are shown in Fig.5 (b). The same sequences of steps are followed for the right hand silhouette to determine the landmark points in the reverse order.

**5.4.** *Major issues*

During these sequential applications of procedures for normalization, two major problems have been observed which may result in incorrect tips and valleys detection. The main reason is improper positioning of the hand during imaging. The analyses of such ideas are described next along with the possible solutions.

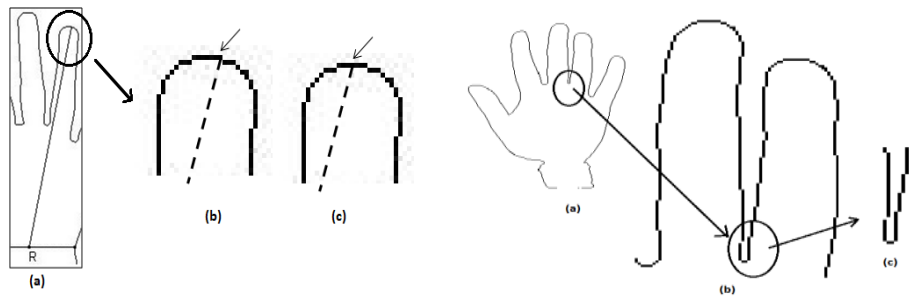

Fig.6. (a) partial tip (b) original tip (c) modified tip.   Fig.7. (a) wrong hand placement (b) and(c) zoomed of (a).

### 5.4.1. *Problem 1:*

The first problem takes place due to faulty positioning of exact tips and valleys. In some cases, it has been observed that mainly thumb is not correctly placed, and that is posed in such a way that the partial shape of thumb is captured in the raw image. In some other cases, it has been found that, for the remaining fingers the highest Euclidean distance from the reference point is not exactly the finger-tip, shown in Fig.6 (b). The reason is that, any finger during image acquisition may be aligned to either left or right side due to pose dissimilarities. This kind of problem has been solved by considering the middle pixel of that end row where the extreme point exists. It implies that several pixels are involved in the finger extreme region. Find those pixels in that end row and locate the midpoint as the finger-tip of the respective finger.

### 5.4.2. *Problem 2:*

Another problem is detected as no spacing between two consecutive fingers during image acquisition. Fig.7 illustrates such wrongly placed hand and its impact on the edge detection. The minimum space for placing the fingers is not maintained. It may cause a discontinuity in the hand contour. If this situation persists, then the exact finger extremes and thus features can't be calculated accurately. This unacceptable hand placement can lead to higher intra-class variations and misclassify a user. So, in peg-free imaging system a little bit of carefulness is necessary to avoid such circumstances. Images captured without specific gaps between fingers should be rejected. The hand should be placed in a relax mode with standard spacing in between fingers, and the hand-device contact should be adequately checked.

Finally, a robust and precise hand image normalization technique is implemented by solving these problems.

## 6.   Feature Extraction

After positioning all tips and valleys, certain consistent geometric characteristics are extracted. Different features are computed such as the length and widths at one-third and two-third position along the length of individual finger. Palm width is measured along a straight line (CS), starting from the valley point (C) of thumb to the other side of the palm(S). The midpoint (M) of this palm width (CS) is considered as the centre point of the normalized hand. The distances from this midpoint (M) to the middle point of every finger baseline (e.g. NM, OM etc.) are measured. The baselines are measured directly by using all the valley points.

In addition to various traditional features, some new features are also included. The length and width of fingers are very familiar features and are measured conventionally. Five new features are the distances from M to the midpoint of all finger baselines. Altogether 26 features are measured from each normalized hand image, illustrated in Fig.8. The feature set includes:

Finger length (1 per finger): 5
Finger width (3 per finger): 15
Length from M to midpoint of finger base lines (1 per finger): 5
Palm width: 1

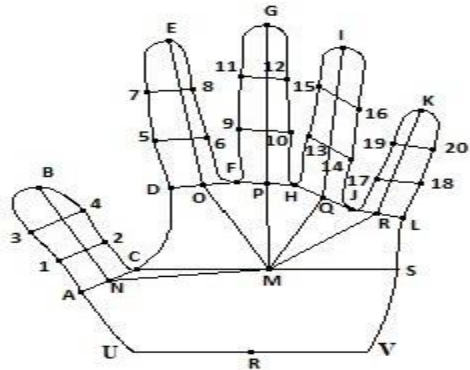

Fig.8. Geometric feature extraction of the left hand.

All the described steps are followed in the same order for the right hand images.

## 7.   Classification Algorithms

The classification algorithms follow simple steps to generate initial probabilities by using the Euclidean distances between enrolled templates and test vector. Details of the algorithms are described as follows.

### 7.1.   *Classifier 1*

This is a modification of the nearest neighbor algorithm, described in Algorithm 1, where three nearest neighbors ($d=3$) are considered. Suppose, $\textbf{\textit{A}}_{n \times x}$ and $\textbf{\textit{B}}_{n \times x}$ are two matrices representing the database of the left and right hand feature templates *i* of *n* enrolled users with *x* (*x*=*26* here) different features, respectively. A feature vector is denoted by *i* that contains *x* unique features. Both types of feature set are employed for the testing purpose. The left and right hand test feature vectors of a claimer *k* are represented by $\boldsymbol{L_k}$ and $\boldsymbol{R_k}$ respectively, whose identity needs to be tested. The sum of feature distances (Euclidean) between the enrolled templates and test vector are added (denoted by $\boldsymbol{S_L}$ and $\boldsymbol{S_R}$), and their average is used as fused sum ($\boldsymbol{S_F}$). The feature level fusion is performed by this method. The minimum sum from each of the $\boldsymbol{S_L}$, $\boldsymbol{S_R}$ and $\boldsymbol{S_F}$ are identified and symbolized as $\boldsymbol{C_L}$, $\boldsymbol{C_R}$ and $\boldsymbol{C_F}$ respectively. From these minimum sum values, corresponding probabilities ($\boldsymbol{P_W}$) are computed, and the decision is made.

### 7.1.1. *Algorithm 1:*

**Input:**
(a) Test feature set: $\mathbf{L_k}$ and $\mathbf{R_k}$
(b) Database: matrix $\mathbf{A_{n \times x}}$ and $\mathbf{B_{n \times x}}$
(c) Number of nearest neighbors: $\mathbf{d}$
**Output:** Class representing maximum Probability

1. Calculate $\mathbf{S_L}$ and $\mathbf{S_R}$ between test vectors ($\mathbf{L_k}$ and $\mathbf{R_k}$) and all feature vectors (i) of $\mathbf{A_{n \times x}}$ and $\mathbf{B_{n \times x}}$ respectively.

   $$S_L(i) = \sum_{j=1}^{x} \sqrt{(A_i^j - L_k^j)^2} \quad \text{and} \quad S_R(i) = \sum_{j=1}^{x} \sqrt{(B_i^j - R_k^j)^2} \ ,$$

   Where, $1 \leq i \leq n$ and $1 \leq j \leq x$

2. The mean of $S_L(i)$ and $S_R(i)$ is calculated.
   $S_F(i) = (S_L(i) + S_R(i))/2$

3. Find the minimum from each of the $S_L(i)$, $S_R(i)$ and $S_F(i)$.
   $C_L(i) = \min(S_L(i))$, $C_R(i) = \min(S_R(i))$ and $C_F(i) = \min(S_F(i))$

4. Calculate the probability as:

   $$P_W(i) = \frac{1}{(d-1)} * \left[ 1 - \frac{C_W(i)}{SS(i)} \right]$$

   W represents the feature types, W=L, R or F

   Where, $SS(i) = C_L(i) + C_R(i) + C_F(i)$ and $\sum P_W(i) = 1$

5. Decision1(i) = argmax($P_L(i)$, $P_R(i)$, $P_F(i)$)
6. End

The minimum values $C_L(i)$, $C_R(i)$ and $C_F(i)$ may represent either same or different feature vector. When the matching is very high, then the sum of distances will be small. As a result, the probabilities ($P_L$, $P_R$ and $P_F$) of matching will be high. Feature level fusion is achieved by averaging the left and right feature vectors, in Step 2. The number of nearest neighbors (*d*) is scalable and can be increased to higher values. To calculate the probability, Step 4 is valid for any higher value of *d*.

## 7.2.   *Classifier 2*

This algorithm is distance threshold dependent, described in Algorithm 2. The database and test feature vectors are same as defined in Classifier1.  First of all, the distances between the test vectors and the enrolled feature vectors are calculated. Then, a distance threshold is considered. The threshold is basically a measure of similarities between the test vector and feature templates. For any pre-specified threshold (*th*), count (*CNT*) the number of features (both left and right hand features) which exist below this threshold, *th*. The average of left count (*CNT_L*) and right count (*CNT_R*) is used to represent the fused count (*CNT_F*). Feature vector *i* with maximum *CNT_F* is represented by $Q_d(i)$. The sum of $CNT_F$ of first *d* neighbor vectors is denoted by *SC*. Here, the probability (*PR*) is calculated as the ratio of the $Q_d$ and *SC* calculated from *d* neighbors.

### 7.2.1.   *Algorithm 2:*

**Input:**
(a) Test feature set: **L_k** and **R_k**
(b) Database: matrix **A_{n×x}** and **B_{n×x}**
(c) Number of nearest neighbors: **d**
**Output:** Class representing maximum Probability

1. For left hand:

   $CL^j(i)=1$ if $\sqrt{(A_i^j - L_k^j)^2} \leq$ th; $CNT_L(i)=\sum_{j=1}^{x} CL^j(i)$

   and for the right hand:

   $CR^j(i)= 1$ if $\sqrt{(B_i^j - R_k^j)^2} \leq$ th; $CNT_R(i)=\sum_{j=1}^{x} CR(i)$

   Where, $1 \leq i \leq n$ and $1 \leq j \leq x$

2. The average count is: $CNT_F(i)=(CNT_L(i)+CNT_R(i))/2$

3. Consider, first d vectors with maximum $CNT_F(i)$:
   $Q_d(i)=max_d(CNT_F(i))$

4. Calculate the probabilities (PR):

   $PR_d(i)=Q_d(i)/SC(i)$, where, $SC(i)=\sum_{h=1}^{d} Q_h(i)$ and $\sum PR_d(i)=1$

5. $Decision2(i)= argmax(PR_1(i),PR_2(i),PR_3(i))$

6. End

The value of *d* neighbors is also scalable. The threshold *th* is user defined. Choosing the value of *d* and *th* is experiment dependent. Other distance metrics (e.g. Hamming, Mahalanobis etc.) can also be applied.  Here, feature level fusion is performed by the average count according to the Step 2. If *W* is the enrolment size per subject (i.e. the number of feature vectors enrolled per user) and if nearest neighbors *d>W*, then, in the best case, first *W* number of highest matching should come from that particular class, and remaining *(d-W)* should be selected from other closely matched class(es).

***Complexity of algorithms:*** As all the unit operations (Matrix addition and subtraction) are done using a 2-D matrix, so the time complexity is $O(n \times x)$, where *n* and *x* are defined as above. Now, *x* is tiny compared to *n* and *x* is a constant, which improves the time complexity to *O(n)*, that means these algorithms can execute in linear time.

### 7.3. *Dempster-Shafer belief functions*

If both the classifers agree on a particular user, then AND rule can be applied. If there is disagreement, depending on BPAs of initial decisions, the final determination is made according to the belief function. Suppose, class A and B are recognized by Classifier1 and Classifier2, respectively with certain probabilities $p_{A1}$ and $p_{B2}$ respectively. Also, find the probabilities of the corresponding classes defined by other classifiers ($p_{A2}$ and $p_{B1}$), where, $p_{A1} \geq p_{B1}$ and $p_{B2} \geq p_{A2}$. The uncertainty of matching is represented by *U*. In an ideal case, for an unauthentic user, the probability (i.e. BPA) of *U* should be the highest and for a genuine user *U* should be the lowest. The BPAs are assigned in Table2.

Table 2. BPAs of classifiers.

| Classifier | Class A | Class B | Uncertain (U) |
|---|---|---|---|
| Classifier 1 | $p_{A1}$ | $p_{B1}$ | $1\text{-}p_{A1}\text{-} p_{B1}$ |
| Classifier 2 | $p_{A2}$ | $p_{B2}$ | $1\text{-}p_{A2}\text{-} p_{B2}$ |

By using $p_i$, the belief functions (*BEL(i)*) are calculated according to Eq.2. Where, the class label, *i=A, B and U*. Finally, the recognized class, RC= argmax *(BEL(i))*.

## 8. Experimental Results

 A random subset of the Bosphorous hand database, created at the Bogazici University [8, 9] has been experimented here. It was actually acquired from the Turkish and French staff members, and students of different universities and their ages vary from 20 to 50. Sufficient intra-class variations in pose and gadget have been considered in the database. HP Scanjet 5300c scanner was used at 45-dpi resolution. Three images per left and right hand of a person were collected at three different scanning sessions within 5-10 minutes of interval. Initial images are with 383×526 pixels and are resized to 200×300 pixels at feature extraction stage. Two sample images per hand of each user are applied for enrolment, and one image is used for testing.

### 8.1. *Experiment Description*

The primary objective of various experiments is to validate the proposed system model and the classification algorithms. Experiments are conducted in the identification and verification mode. The particulars of all experiments are described below.

*Experiment 1:*

In the first experiment (Exp.1) altogether three classes, the minimum sum from each of the left ($C_L$), right ($C_R$) and fused ($C_F$) feature vectors are considered (step 3) to calculate the initial probabilities (step 4) and decisions are taken accordingly (step 5) by the Classifier1. The second classifier also considers three classes from the maximum fused feature count ($CNT_F$) for the calculation of the necessary probabilities (step 4) and decisions are made (step 5), respectively. Each of the classes recognized by Classifier1 is compared to each class of Classifier2 to find if there is any matching in their decisions. The last decision is accomplished by the maximum belief function i.e. according to Eq. 3.

*Experiment 2:*

In Experiment2, (Exp.2) the same method is followed as done in Classifier1, described in Exp.1. Here also the minimum sum from each of the left, right and fused vectors is used for BPA calculation. Out of the three classes, two or all of them may represent the same class. If so, then the probabilities of the same class are added, and the class with maximum probability is considered as Decision1. Again, three classes one from each of the three feature vector count ($CNT_L$, $CNT_R$, and $CNT_F$) is considered to estimate initial probabilities according to the following formula:  $PR_d(i)= Q_d(i)/x$                    (6)
where, $x$ is number of features as defined above. The classes with maximum probabilities are defined as Decision2 by the Classifier2. Now, based on the Decision1 and Decision2, the final the decision is made according to the maximum probability, using Eq. 3.

*Experiment 3:*

In the last experiment (Exp.3), three classes all from fused feature vectors ($C_F$ and $CNT_F$) are used to calculate the initial probabilities. This has been accomplished by considering fused minimum distance sum ($S_F$) by Classifier1 (step 3) and maximum fused feature count ($CNT_F$) by Classifier2 (step 3).  To calculate initial probabilities by the Classifier1, step 4 has been modified as:

$$P_F(i) = \frac{1}{(d-1)} * \left[ 1 - \frac{C_F(i)}{SS(i)} \right], where\ SS(i) = \sum_{i=1}^{d} C_F(i)\ .$$     (7)

Classifier2 remains unaltered. Each class from both of the classifiers are compared to find if there is any matching in initial decisions. Final decision is considered according to the Eq.3. The differences among the experiments exist in the first level of decision making.  In Exp.2, Classifer2 uses a different formula (Eq. 6) for BPA calculation, and the sum rule is applied. In Exp.3, Classifier 1 applies a modified step by step 4 (Eq.7). But, the final decision making stage is same for all the experiments.

### 8.2. *Identification*

In identification mode, a claimer feature set is compared against the templates of all subjects stored in the database. To establish a user identity, respective feature template is matched (one-to-many) with all the templates. An authentic user is recognized correctly when the matching probability with the test feature set is the maximum. In the first test of identification, the results are obtained without feature level and decision level fusion. The second experiment is performed based on the feature level fusion only. Classifier 1 is based on the minimum distance classifier and does not vary with respect to the threshold; so it produces 98% average accuracy. Classifier2 yields minimum 84% and maximum 98.5% accuracy at 0.3 and 1.2 threshold values, respectively. The results of these two experiments are given in Table 3. In the next set of experiments (given in Table 4), in addition to feature level fusion, belief function is used for final decision making and the performance improvement is very significant, particularly in Exp.2 and Exp.3.

Table 3. Identification results using feature level fusion.

| Threshold | 0.7 | 1.2 | 1.6 | 2.3 | 2.5 |
|---|---|---|---|---|---|
| Without fusion | 94 | 97 | 96 | 96.5 | 96 |
| Feature level fusion | 97.7 | 98.5 | 98 | 97.7 | 97 |

Table 4. Identification results using decision level fusion.

| Threshold | 0.3 | 0.7 | 1.2 | 1.6 | 2.3 | 2.5 |
|---|---|---|---|---|---|---|
| Exp. 1 | 96.5 | 97 | 98.5 | 98.5 | 97.5 | 97 |
| Exp. 2 | 98 | 99.5 | 98.5 | 97.5 | 97 | 96.5 |
| Exp. 3 | 94 | 98 | 99 | 98 | 98.5 | 98 |

The results in Table3 and Table4 imply that for a particular threshold, the average performance without decision fusion is approximately 1% to 4% lesser than using fusion. Exp.2 and Exp.3 provide the best results at 0.7 and 1.2 distance threshold respectively.

Table 5. Performance with diiferent population sizes with respect to a threshold.

| Threshold 0.7 | | | | | Threshold 1.2 | | | | |
|---|---|---|---|---|---|---|---|---|---|
| Population | 50 | 100 | 150 | 201 | Population | 50 | 100 | 150 | 201 |
| Exp. 1 | 98 | 97 | 97.3 | 97 | Exp. 1 | 98 | 97 | 97.3 | 98 .5 |
| Exp. 2 | 98 | 99 | 99.3 | 99.5 | Exp. 2 | 98 | 97 | 98 | 98.5 |
| Exp. 3 | 96 | 96 | 97.3 | 98 | Exp. 3 | 98 | 96 | 98.7 | 99 |

The performance with varying population sizes is also computed at two different threshold values 0.7 and 1.2, respectively. All the experiments are producing very close identification results particularly at threshold 1.2, shown in Table5.

### 8.3.  *Verification*

In the verification mode, for a claimed template feature differences against the given threshold of the claimer and the enrolled users are calculated. If the differences are within the threshold then the user is accepted, or else rejected. The performance is measured in terms of the False Accept Rate (FAR) and False Reject Rate (FRR). The Genuine Accept Rate (GAR) = 1-FRR.

The experiments are carried out with 201 genuine users, and two images are used for enrolment and one image used for verification. Total 3×201×201=121203 trials are tried out, of which 201 are genuine, and 2×200×201=80400 are imposters. Therefore,

*FAR (%) = (wrongly accepted users/80400) ×100*

*GAR (%) = (correctly accepted uses/201) ×100*

Table 6.  Verification performances of proposed work.

| Proposed work | GAR(%) at 0.00125% FAR | FAR(%) at 99.5% GAR |
|---|---|---|
| Exp. 1 | 77.62 | 3.96 |
| Exp. 2 | 69.65 | 0.625 |
| Exp. 3 | 72.64 | 3.13 |

Minimum FAR obtained for all experiments is 0.00125%, and maximum GAR is 99.5% are obtained from different experiments. Table6 shows the results at the specified FAR and GAR. Exp.1 and Exp.2 yield the best result at the lowest FAR and highest GAR, respectively. Exp.3 produces approximately the in-between results of Exp.1 and Exp.2 in both cases. Exp.2 produces below 1% FAR at the maximum GAR, which is very significant improvement over previously reported contributions.

### 8.4.  *Performance Comparisons*

A straightforward performance comparison with some previous works, both unimodal and multimodal hand geometry is given in Table7. Though the methods at different stages, feature characteristics, classification algorithms, database with population size, image quality, and other parameters are not similar in those works, direct comparison is shown for the completeness of the study made in this paper.

Table 7.  Comparisons of Identification and Verification.

| Identification | | | Verification | | |
|---|---|---|---|---|---|
| [Ref.]Authors, Year (%) | Users (images) | Result in | [Ref.]Authors, Year | Users (images) | Result in (%) |
| [14]Sanchez *et al.*,2000 | 20 (10) | 97 | [5]Jain *et al.*,2003 | 50(5) | FAR :0.03, FRR=1.78 |
| [11]Zhang *et al.*, 2006 | 100 (10) | 98 | [6]Jain *et al.*,2005 | 100( 5) | FAR :0.1, FRR=1.4 |
| [9]Yörük *et al.*, 2006 | 458(3) | 97.21 | [21]Alonso *et al.*, 2007 | 106(10) | FAR:0.22 ,FRR:0.15 |
| [20]Zhou *et al.*, 2010 | 86(10) | 99.37 | [9]Yörük *et al.*, 2006 | 458(3) | GAR=98.21(at ERR) |
| [27]Le *et al.*, 2011 | 20(80) | 98.75 | [19]Azad *et al.*, 2009 | 200(3) | FAR:1.33,FRR:0.5 |
| **Proposed , 2015** | **201(3)** | **99.5** | **Proposed, 2015** | **201(3)** | **FAR: 0.625, FRR:0.5** |

In the identification, our contribution is satisfactory with the magnitude of population. In verification, though Ref. 19, produced same GAR as ours, but the FAR they reported is more than double compared to our Exp.2. So, in both identification and verification this work produces acceptable results in comparison with the existing state of the arts. The Receiver Operating Characteristic Curve (ROC) is plotted for every Experiment, in Fig.9.

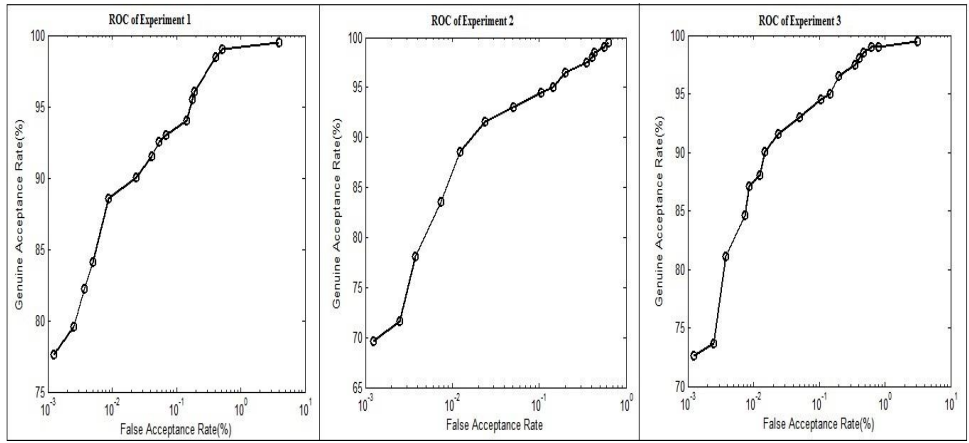

Fig.9. ROC Curve of the three Experiments.

## 9. Conclusion

A new hand biometric based identification system using fusion is presented with the implementation details. Discussions on some major problems and their respective solutions are proposed for more accurate feature measurements and identification. The main advantage of this fusion strategy is the lower implementation cost as it deals with only single biometric mode and thus no extra sensor is required. As only some simple geometric features are introduced, it also reduces the feature extraction complexities. Feature level fusion is carried out to represent robust identity of a person. Decision fusion using belief function produces excellent results. The experimental results imply that, this system provides an acceptable identification and verification performance for security based applications with improved accuracy. Our next objective is to substantiate a more accurate hand template generation at the normalization stage. The performance of the system can be experienced and verified by using some other classifiers (SVM) at different fusion level (rank level) as a future enhancement. The stability of classification algorithms can be further improved, and the population size can be increased to raise the applicability of this biometric system in various domains.

**Acknowledgment**

The authors would like to thank the Editors and anonymous reviewers for their valuable comments and Prof. B. Sankur of Bogazici University for providing the hand image database used in this paper. The authors are thankful to the department of Computer Science and Engineering, Jadavpur University, India for providing the necessary infrastructure to conduct the experiments related to this work.